\documentclass[conference, 11pt]{IEEEtran}
\IEEEoverridecommandlockouts
\usepackage{cite}
\usepackage{amsmath,amssymb,amsfonts}
\usepackage{graphicx}
\usepackage{textcomp}
\usepackage{xcolor}
\usepackage{url}
\usepackage{graphicx}
\usepackage{algorithm}
\usepackage{algpseudocode}
\usepackage{float}
\usepackage{geometry}
\usepackage{multirow}
\usepackage{booktabs} 
\def\BibTeX{{\rm B\kern-.05em{\sc i\kern-.025em b}\kern-.08em
    T\kern-.1667em\lower.7ex\hbox{E}\kern-.125emX}}
\begin{document}

\title{{Reinforcement of Explainability of ChatGPT Prompts by Embedding Breast Cancer Self-Screening Rules into AI Responses}
}

\author{\IEEEauthorblockN{1\textsuperscript{st} Yousef Khan}
\IEEEauthorblockA{\textit{Sano Centre for Computational Medicine} \\
Krakow, Poland \\
y.khan@sanoscience.org}
\and
\IEEEauthorblockN{2\textsuperscript{nd} Ahmed Abdeen Hamed}
\IEEEauthorblockA{\textit{SSIE Department, CASCI Laboratory} \\
\textit{Binghamton University}\\
New York, USA \\ Corresponding: ahamed1@binghamton.edu*}
}
\maketitle

\begin{abstract}
    \textbf{Background:} Addressing the global challenge of breast cancer, this research explores the fusion of generative AI, focusing on ChatGPT 3.5 turbo model, and the intricacies of breast cancer risk assessment. \textbf{Objective:} The research aims to evaluate ChatGPT's reasoning capabilities, emphasizing its potential to process rules and provide explanations for screening recommendations. The study seeks to bridge the technology gap between intelligent machines and clinicians by demonstrating ChatGPT's unique proficiency in natural language reasoning. \textbf{Methods:} The methodology employs a supervised prompt-engineering approach to enforce detailed explanations for ChatGPT's recommendations. Synthetic use cases, generated algorithmically, serve as the testing ground for the encoded rules, evaluating the model's processing prowess. \textbf{Conclusion:} Findings highlight ChatGPT's promising capacity in processing rules comparable to Expert System Shells, with a focus on natural language reasoning. The research introduces the concept of reinforcement explainability, showcasing its potential in elucidating outcomes and facilitating user-friendly interfaces for breast cancer risk assessment.
\end{abstract}


\section{Introduction}
Generative Artificial Intelligence (AI) represents a departure from traditional rule-driven systems, offering the capacity to generate contextually relevant responses and insights. Unlike conventional AI approaches, generative models possess the flexibility to learn patterns, adapt to diverse datasets, and generate novel outputs. New Generative AI tools come with a lot of skepticism due to issues of lack of credibility and transparency, making it prevalent that these new tools come with explainable algorithms to help give reassurance in results produced from LLMs \cite{hamed2023arXiv,hamed2023challenging,HAMED2024}. Through analysis, such as medical literature, Large Language Models (LLMs) can offer doctors valuable insights into uncommon conditions, propose potential treatment approaches, and even forecast patient outcomes \cite{sysmaticlitreview,TechnischeUniversitatDresden2021}. Recent advancements in AI have even shown significant improvements in detecting skin cancer and lymphoma from medical imaging \cite{Amann2020,BMJ2019}. AI-driven methodologies have demonstrated high performance in oncology, highlighting AI's potential for widespread application in healthcare \cite{BMJ2019} such as building decision making systems \cite{10.1093/jamia/ocac121,mittal2020no} and accelerate diagnosis \cite{Wang:2022:0368-492X:1173,liang2022integrating}. These capabilities highlight the transformative potential of generative AI in the healthcare industry, in areas like disease diagnosis, treatment, and patient education \cite{doi:10.1148/radiol.223312,Kung2023,Patel2023,Singhal2023,Wang2023HealthcareLLM}. A prominent exemplar of generative AI, ChatGPT \cite{ChatGPT}, leverages a vast array of pre-existing knowledge to understand and respond to user inputs with remarkable coherence and context-awareness. This paradigm shift in AI methodology opens doors to dynamic, adaptive decision-making systems that can navigate nuanced scenarios with a level of sophistication previously unparalleled.

Though ample evidence highlights the risks associated with searching for health information online \cite{10.1093/heapro/dag409,10.1145/1629096.1629101,10.1007/978-3-319-16354-3_62}, individuals persist in turning to online platforms for such information \cite{PewResearch2011}. While research has explored the efficacy of traditional search engines in delivering health-related answers \cite{10.1007/978-3-319-16354-3_62}, in comparison, investigation into LLMs is absent. While health chatbots based on preLLM technology have been introduced \cite{10.1145/3589959}, there remains a gap in comprehensively evaluating and understanding chatbot solutions utilizing the latest methods in generative LLMs. This paper aims to further explore this matter by examining two primary experimental conditions and focusing on the particular widely-used LLM called ChatGPT, which is currently relied upon by end-users.

Integral to this research is the concept of prompt engineering, a method by which diction and arrangement of examples within a prompt significantly impact the behavior of the model, its guided-to process and ability to understand specific information \cite{fantastically,prompt}, this technique that differs from fine-tuning as it doesn't alter the pretrained model's weights for downstream tasks \cite{10.1145/3560815}. Previous research has demonstrated the significant influence of prompt engineering on the efficacy of LLMs like ChatGPT, commonly employed to facilitate few-shot or zero-shot learning tasks \cite{NEURIPS2020_1457c0d6,kojima2022large}, diminishes the necessity for model fine-tuning and dependence on supervised labels as it refines the behaviour of LLMs to improve its performance. 

In the context of breast cancer risk assessment, supervised prompt engineering involves the systematic feeding of rules one at a time to train ChatGPT. This supervised approach ensures that the model accurately processes and makes decisions based on the encoded rules, simulating an expert system shell. The objective is not only to elicit accurate recommendations from ChatGPT but also to force detailed explanations of the underlying rules, providing transparency and interpretability to the decision-making process. The subsequent sections outline the structured process of supervised prompt engineering, shedding light on its pivotal role in training ChatGPT for effective breast cancer risk assessment.

Breast cancer is a pervasive global health challenge, demanding innovative approaches for early detection and accurate risk assessment. In the pursuit of advancing these efforts, this research navigates the crossroads of generative AI, with a specific focus on ChatGPT, and the intricacies of breast cancer risk assessment. Here we investigate and scrutinize ChatGPT's capacity for reasoning, particularly in its ability to comprehend rules extracted from the guidelines established by the American Cancer Society (ACS) \cite{ACS2024}. But most importantly, it's ability to explain and justify the utilization of rules to make a screening recommendation. As large language models become more prevalent for information seeking tasks, especially those influencing critical health-related decisions like treatment options, it is crucial to enhance our comprehension of these models, encompassing aspects such as the accuracy of their responses \cite{koopman-zuccon-2023-dr}.Our investigations have demonstrated a promising capability comparable in ways to expert system shells in a way rules are processed and recommendations are explained. More interestingly, ChatGPT reasoning engine enables the processing of rules in a natural language, making it possible for domain experts and clinicians to overcome the difficulties of programming Expert System Shells (ESS), which may bridge the technology gap between intelligent machines and clinicians.

While conventional rule-based systems have been the norm, the integration of generative AI, as demonstrated by ChatGPT presents a new promising avenue that may enhance the early detection of breast cancer. AI's ability to analyze large datasets accurately and rapidly is proving transformative in early cancer detection and treatment (Nature, 2023)\cite{Foronda2020}. The evolution from traditional methods to AI-powered diagnostic tools has not only improved diagnostic accuracy but also enhanced early disease detection, crucial for effective treatment \cite{BMJ2019}.In the next sections, we present our generative AI on encoding knowledge embedded in the ACS guidelines as rule which we use to ``train'' the ChatGPT prompt-engineering engine. Using the different features presented in each of the guidelines, we generated 50 use cases to test these rules and determine the performance of the ChatGPT engine in processing the rule and make recommendations. Another important aspect we present in this paper is what we call ``Explainability Enforcement'' where we inject certain ``commands'' to trigger ChatGPT to explain its prompt response, and we present the results.

\section{Data}
The data used in this publication is from two different sources of different types: (1) Rules extracted from the American Cancer Society website \cite{ACS2024}; (2) Algorithmically generated use cases which we describe in the methods section. 

\section{Methods}
The foundation of this research rests at the intersection of generative AI, with a specific focus on ChatGPT, and breast cancer risk assessment. Therefore, the Methods paper are designed around the premise of (1) whether we prompt ChatGPT using clinical rules and trigger responses similar in the same way Expert System Shells made in the past, (2) whether we can enforce the default behavior of ChatGPT to provide detailed explanations for the  recommendations it makes. In the remaining part of the section, we show these two steps on two phases: \textbf{Phase I:} how to encode the breast cancer self-screening rules using (If/Else) style prompt-engineer ChatGPT in a supervised fashioned. The reason for describing this phase as ``supervised'' is because the true purpose of this phase is train ChatGPT using the set of encoded rules, without expectation of any answers, other than a confirmation of the rules being entered; \textbf{Phase II:} is how the rules are tested using use cases we designed algorithmically to test the rules using a set of configurations. 

\subsection{Phase I: Rule Extraction, Rule Encoding, and Prompt-Engineering}
In the initial phase, we undertook the manual extraction of rules from the guidelines outlined by the American Cancer Society (ACS) \cite{ACS2024}. These rules, organized as if-then statements, were identified and documented for the purpose of self-screening for breast cancer. Following the manual extraction, the identified rules were encoded programmatically to facilitate integration with ChatGPT API prompt-engineering. We encoded the rules one at a time, ensuring accurate representation and adherence to the guidelines provided by the ACS. This encoding process aimed to enhance the model's understanding of the rules, enabling it to effectively apply them during subsequent analyses.
\begin{figure}[H]   
    \centering
    \includegraphics[width=1\columnwidth]{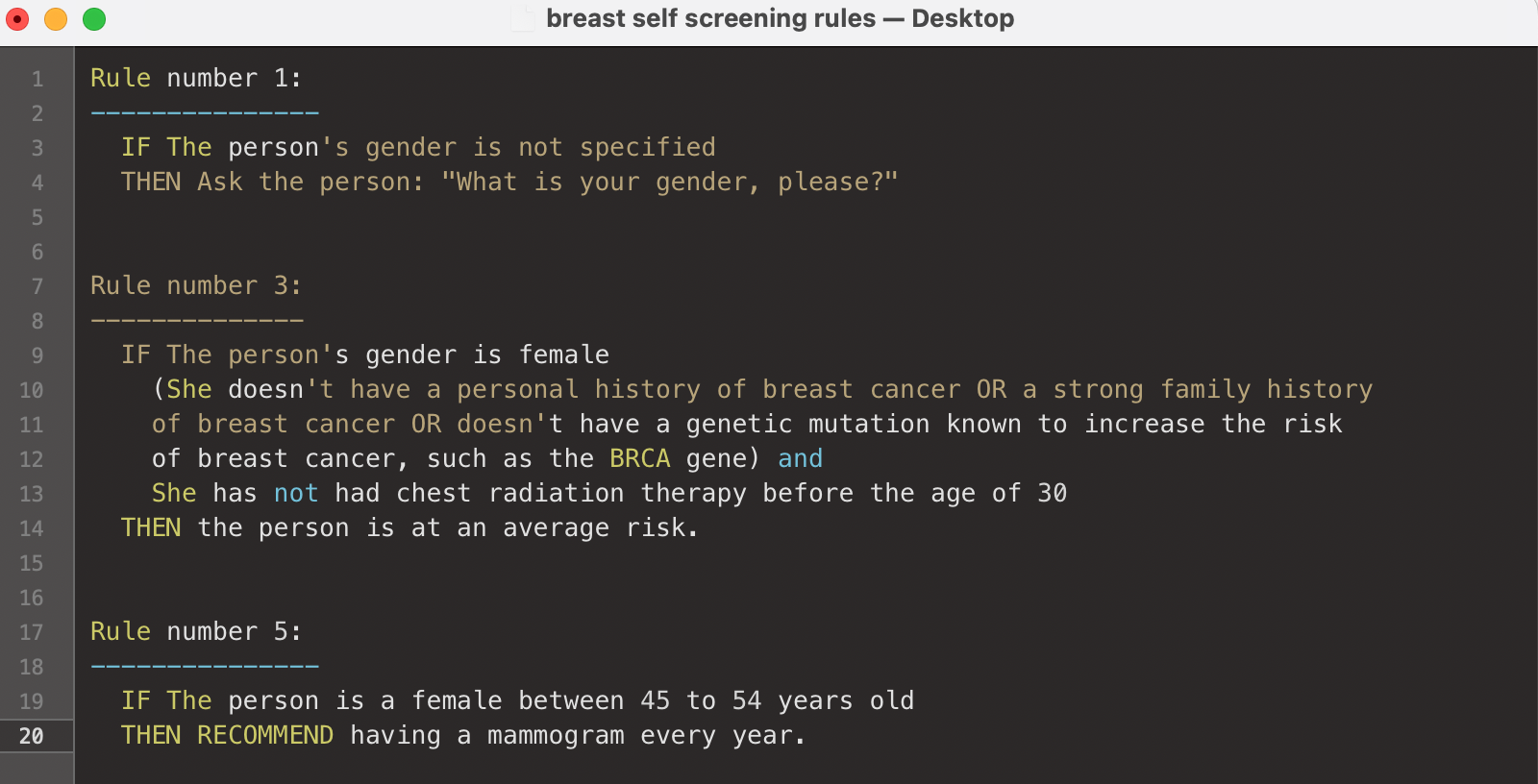}
    \caption{Sample of Rules Extracted.}
    \label{fig:rules3to6}
\end{figure} 
\begin{algorithm}[H]
\caption{Supervised Prompt Engineering of Rules} \label{spr-pe}
\begin{algorithmic}[1]
    \State \textbf{input:} System role is an Expert System Shell
    \State \textbf{request content:}
    \State Assign a session\_id for retrieval \Comment{Create a session for all the rules}
    \For{Each rule in the set of rules}
        \State Encode the rule in the content
        \State Request ChatGPT to assign a rule\_ID
        \State Request confirmation from ChatGPT
    \EndFor
\end{algorithmic}
\end{algorithm}

\subsection{Phase II: Testing the rules using Generated Use-Cases}
To evaluate the performance of ChatGPT in reasoning with the encoded rules, we generated 50 synthetic use cases of structured and unstructured use cases also to determine if ChatGPT can comprehend one type of prompt more accurately than another or if they are evaluated at the same success rate. These use cases included information such as an individual's medical history, gender, and risk factors for breast cancer. The synthetic data set serves as a controlled environment to systematically test the capabilities of ChatGPT in applying the encoded rules to diverse scenarios. The process of generating 50 synthetic use cases to test the rules which were extracted from the data collection section described above were observed using a python algorithm developed based off known correlations between those who have risk factors related to breast cancer vs those who are less impacted extracted from the ACS. This algorithm performs the following precise steps: (1)  Enumerate risk factors such as known BRCA1/BRCA2 gene mutations, family history, radiation therapy, specific syndromes, and personal history of breast cancer, among others; (2) The algorithm goes on to systematically create diverse synthetic use cases with random combinations of gender, age, and breast cancer risk factors. These synthetic use cases are instrumental in evaluating the rules' efficacy in identifying potential breast cancer risks.; (3) the actual encoding of the script to perform the task of generating synthetic use cases can be captured algorithmically in Algorithm \ref{algo:gsuc}: 

\begin{algorithm}
\caption{Algorithmic Generation Use Cases Algorithm} \label{algo:gsuc}
\begin{algorithmic}[1] 
    \State \textbf{Require:} List of 50 synthetic use cases
    \State \textbf{Require:} List of screening features contains (gender, age, medical history, and risk factors related to breast cancer)
    \For{each useCase in range of 50} \Comment{ -- Data Generation Step --}
        \For{each config\_param in screening features}
            \State Randomly choose gender
            \State Randomly choose age between 16 and 90
            \State Randomly choose 1-4 risk factors
            \State Create a background history for the person
            \State Append the use case to the list
        \EndFor
    \EndFor
\end{algorithmic}
\end{algorithm}

\subsection{Rule Testing using Explainable Prompt-Engineering}
Here we describe the process of feeding the rules which were extracted from the data collection section described above. Using what we call ``Supervised prompt-engieering'' which we instructed ChatGPT engine to perform the following precise steps: (1) instructed ChatGPT to act as an expert system shell and start a session specifically for this project, which can be referred to later by an ID. This serves the purpose of making sure we have control over the input to the engine vs using the default behavior of the main ChatGPT engine; (2)Supervised-prompt where we encode the rules one at a time, to train the ChatGPT engine to process and made a decision based on a given use case to also be entered; (3) the expectation of this supervised prompt is to force explanation of the recommendations made by the rules upon firing which is the premise of this work; (4) the actual encoding of the prompt performing the task of supervised prompt-engineering can be captured algorithmically in Algorithm \ref{pr-us}:
\begin{algorithm}[H]
\caption{Prompt Engineering of Use Cases} \label{pr-us}
\begin{algorithmic}[1]
    \State \textbf{input:} System role is an Expert System Shell
    \State \textbf{request content:}
    \State Assign a session\_id for retrieval \Comment{ -- Supervised Step --}
    \For{Each use case in the list of cases} \Comment{ -- Testing Step --}
        \State Prompt the case
        \State Request a recommendation
        \State Request the explanation of which rule(s) are triggered
    \EndFor
\end{algorithmic}
\end{algorithm}

This algorithm systematically evaluated each use case against the encoded rules, determining their effectiveness in identifying potential risks for breast cancer. The algorithm considered various factors such as the accuracy of rule application and the model's ability to correctly interpret and respond to each use case.

\section{Results}
Our experiment measures the performance for two different types of experiments: (1) testing against a structured use case prompts, (2) unstructured use cases, written in natural languages and formatted as plain-text. 

\begin{table*}
    \centering
    \caption{Outcome of Use Case Evaluation:}
    \label{table:rulesstats}    
    \begin{tabular}{@{}ccc|ccc@{}}
        \toprule
        \textbf{Use Case Type} & \textbf{\# of Correct} & \textbf{\# of Incorrect} & \textbf{1-Rule Triggered} & \textbf{N-Rule(s) Triggered} \\
        \midrule
        Structured Use Case  & 47 & 3 & 47 & 3 \\
        Unstructured Use Case  & 41 & 9 & 46 & 4 \\
        \bottomrule
    \end{tabular}
\end{table*}

\begin{figure}
    \centering
    \includegraphics[width=.95\columnwidth]{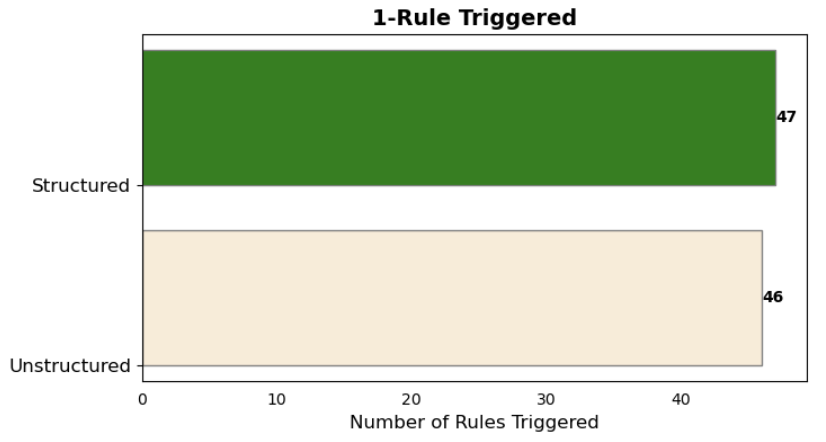}
    \includegraphics[width=.95\columnwidth]{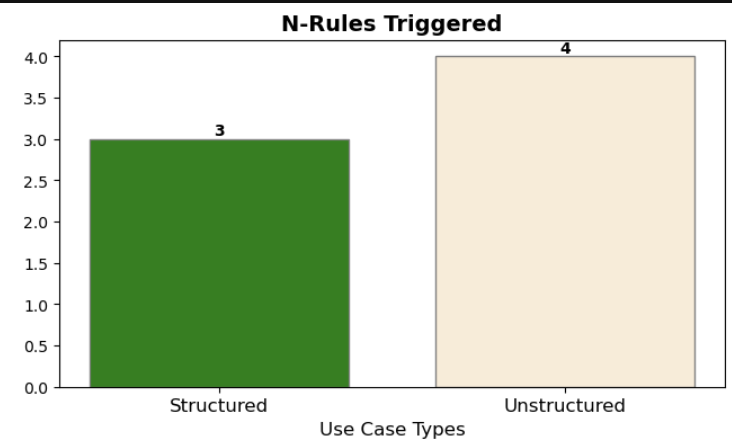}
    \caption{Comparing Results Between Use Case Type}
    \label{fig:triggeredComparison}
\end{figure}

\subsection{Structured Use Case Analysis}
The results in Figure \ref{fig:triggeredComparison} reveal that, for the 50 structured use cases, there were a total of 47 cases where only 1 rule was triggered, while 3 cases had zero rules triggered (seen in the N-Rule(s) Triggered). Regarding the recommendations, 47 cases produced correct recommendations, while 3 cases received incorrect recommendations as shown in Table \ref{table:rulesstats}. It is noteworthy to mention that the 3 cases with incorrect recommendations does not correlate at all with the 3 cases that had 0 rules triggered. In other words, the absence of a rule or rules being triggered does not guarantee incorrect recommendations in these instances. Out of the 50 rules triggered, the distribution indicates a prevalence of single rule triggers across structured use cases.

\subsection{Unstructured Use Case Analysis}
For the 50 unstructured use cases, we observe in Table \ref{table:rulesstats} that 46 cases had only 1 rule triggered, and 4 cases had multiple rules triggered (N-Rule(s) Triggered). Interestingly, in the unstructured cases, there were no instances where 0 rules were triggered for a use case. Regarding the recommendations, 41 cases had correct recommendations, while 9 cases received incorrect recommendations. The absence of cases with zero rule triggers in unstructured use cases suggests a consistent application of rules, even if only a single rule is triggered. This could indicate a higher level of complexity in rule application for unstructured use cases.

\section{Discussion}

\begin{figure}
    \centering
    \includegraphics[width=45mm,scale=0.45]{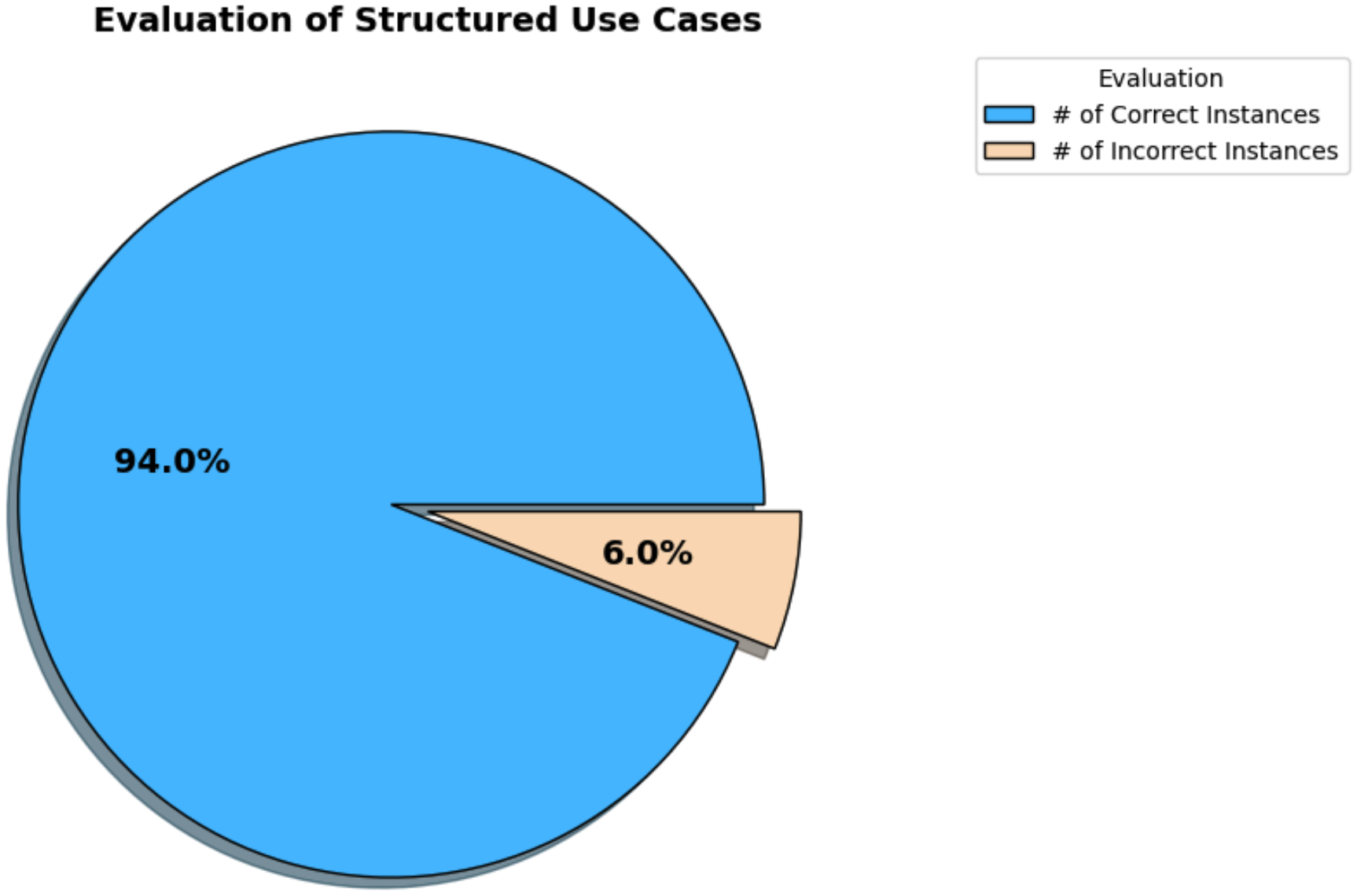}
    
    \includegraphics[width=45mm,scale=0.45]{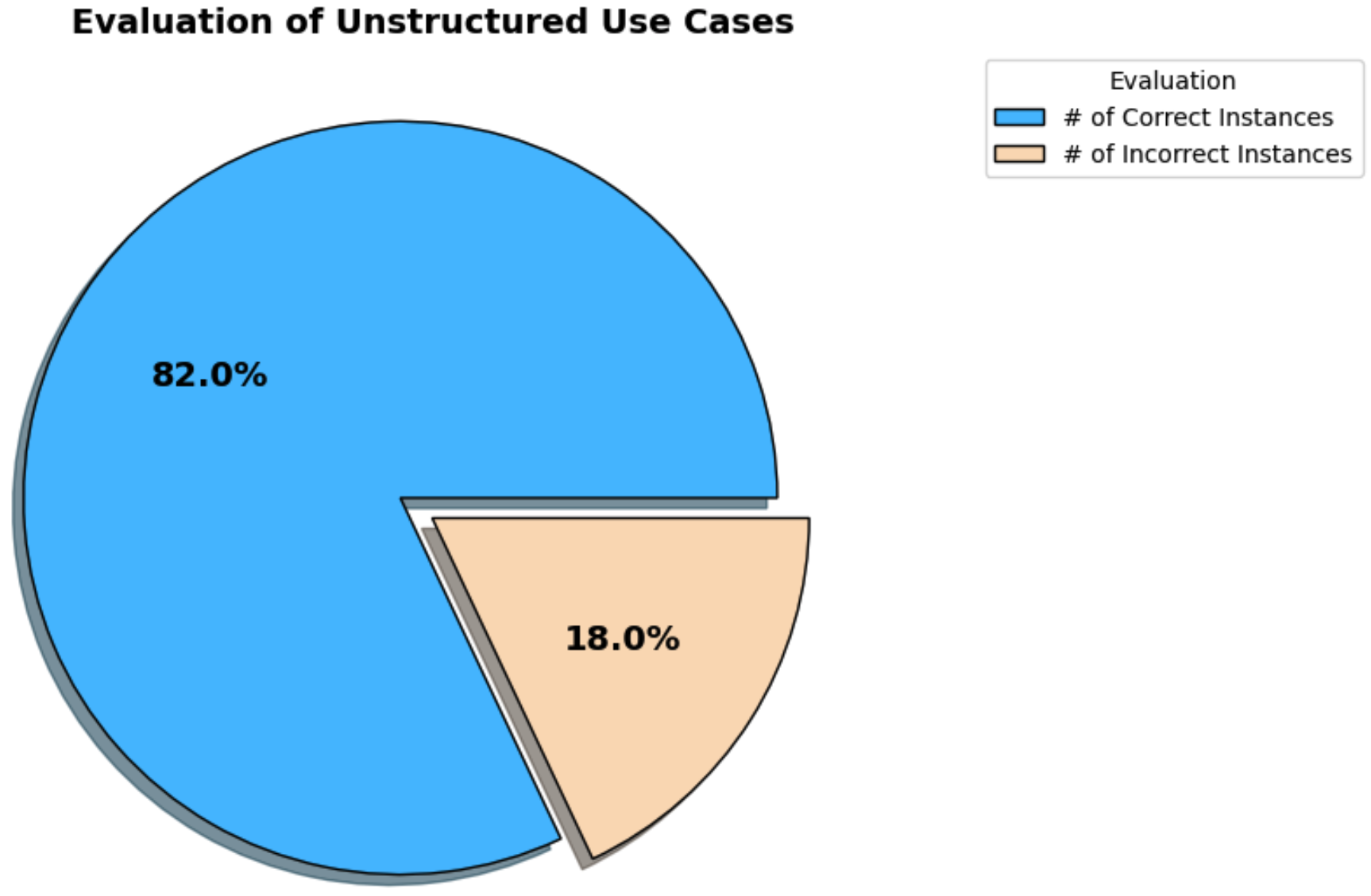}
    \caption{Comparing Results Between Use Case Type}
    \label{fig:pichart}
\end{figure}

The outcomes of our use case evaluation uncover intriguing disparities in ChatGPT's responses to structured and unstructured scenarios. In structured use cases, where a predefined framework guides the evaluation, ChatGPT demonstrated an impressive accuracy of 94\%, primarily attributed to the consistent triggering of a single rule in 47 out of 50 cases, as illustrated in Figure \ref{fig:pichart}. This underscores the efficacy of applied rules in generating accurate recommendations within a structured context. However, the revelation of three instances with incorrect recommendations in structured cases emphasizes the necessity for a nuanced understanding of contextual intricacies.

In contrast, the evaluation of unstructured use cases, marked by a lack of predefined structure, yielded a slightly lower accuracy of 82\%, highlighting the inherent challenges in handling natural language within unstructured scenarios seen in Figure \ref{fig:pichart}. Notably, the absence of instances with zero rules triggered in unstructured cases suggests a consistent application of rules even in cases with multiple rules triggered. This disparity in accuracy between structured and unstructured scenarios underscores the potential impact of structure on the effectiveness of the rule-based system, indicating that a predefined structure can enhance recommendation accuracy in specific contexts.

Examining the rules triggered provides further insights. In structured cases, the consistent triggering of a single rule suggests a well-tailored rule set that aligns with the structured nature of the use cases. However, the cases with incorrect recommendations prompt a closer examination of the rules triggered to understand potential shortcomings or misinterpretations. In unstructured cases, the absence of instances with zero rules triggered indicates the robustness of the rule-based system in consistently applying rules. As this work remains a work-in-progress, further exploration into the nature of incorrect recommendations and a detailed analysis of rule applications in both structured and unstructured cases will be pivotal for refining ChatGPT's performance across diverse use case scenarios. While initial results provide valuable insights into ChatGPT's responsiveness, additional tests are warranted to strengthen our initial results, considering the potential variation in how each model weighs different features.

\section{Conclusions and Future Directions}
Our work demonstrates the effectiveness of training ChatGPT with reinforcement explainability to process rules and provide clear explanations. The implemented Clinical Decision Support System (CDSS) successfully encodes publicly available knowledge, making it accessible for clinicians and user-friendly for the general public. The democratization of knowledge and rules using ChatGPT holds promise for enhancing accessibility and usability in healthcare decision support systems.

Looking forward, we envision a future where explainability is prioritized, and large repositories of universal knowledge are incorporated into ChatGPT engines through collaboration with domain experts. This collaborative effort aims to refine ChatGPT's capabilities for providing accurate, contextually relevant, and explainable recommendations. Additionally, our ongoing research explores the impact of "prompt engineering" roles, such as clinicians, female patients, and oncologists, on the system's accuracy, offering insights into tailoring the CDSS to diverse user needs. We continue to extend the system's capabilities by incorporating additional rules, protocols, and guidelines to handle a broader range of scenarios and address potential conflicts for further advancements in AI-driven decision support systems in healthcare.

\section*{Acknowledgment}
This publication is created as part of the Ministry of Science and Higher Education’s initiative to support the activities of Excellence Centers established in Poland under the Horizon 2020 program based on the agreement No MEiN/2023/DIR/3796.

\bibliographystyle{IEEEtran}

\bibliography{mybib}

\vspace{12pt}

\end{document}